\documentclass[acmtog,nonacm]{acmart}
\PassOptionsToPackage{x11names,table,dvipsnames}{xcolor}
\acmSubmissionID{papers$\_$397}

\usepackage[ruled]{algorithm2e} %
\usepackage{makecell}

\SetAlFnt{\small}
\SetAlCapFnt{\small}
\SetAlCapNameFnt{\small}
\SetAlCapHSkip{0pt}

\acmJournal{TOG}

\newcommand{\ourwork}{Compact~NGP}

\newcommand{\trainingslowdown}{$1.2$--${2.6\times}$}

\newcommand{\bigo}{\mathcal{O}}
\newcommand{\featuredim}{F}
\newcommand{\levels}{L}
\newcommand{\vertex}{v}
\newcommand{\vertexvec}{\mathbf{\vertex}}
\newcommand{\applicationdim}{d}
\newcommand{\codebookfeat}{D_f}
\newcommand{\codebookindex}{D_c}
\newcommand{\codebookindexwidehat}{\widehat{D}_c}
\newcommand{\codebookfeatsize}{N_f}
\newcommand{\codebookindexsize}{N_c}
\newcommand{\codebookindexcoverage}{N_p}
\newcommand{\minresolution}{N_\mathrm{min}}
\newcommand{\maxresolution}{N_\mathrm{max}}

\newcommand{\hyponet}[1]{\color{red}\mathbf{\Phi}}

\newcommand{\posteg}[1]{}

\newcommand{\pos}{\mathbf{x}}
\newcommand{\R}{\mathbb{R}}
\newcommand{\Z}{\mathbb{Z}}
\newcommand{\N}{\mathbb{N}}

\def\equationautorefname~#1\null{%
  Equation~(#1)\null
}

\usepackage{layouts}
\makeatletter
\newcommand{\layoutdetails}{%
\begin{tabular}{ll}
 \texttt{\textbackslash{textwidth}} & \printinunitsof{in}\prntlen{\textwidth}, \printinunitsof{pts}\prntlen{\textwidth} \\
\texttt{\textbackslash{linewidth}} & \printinunitsof{in}\prntlen{\linewidth}, \printinunitsof{pts}\prntlen{\linewidth} \\
Main text font &  \f@size pt \f@family \\
\sffamily \small Caption text font &  \sffamily \small \f@size pt \f@family \\
\end{tabular}%
}
\makeatother

\usepackage{tabularx}
\usepackage{booktabs}
\usepackage{tikz}
\usepackage{wrapfig}
\usepackage[binary-units]{siunitx}
\usepackage{multirow}
\usepackage{nicefrac}
\usepackage{float}

\definecolor{orange_cubic}{rgb}{.9765, .5887, .3569}
\definecolor{purple_cubic}{rgb}{.4706, 0, .5216}
\definecolor{green_cubic}{rgb}{.28603, .81178, .5008}

\definecolor{grayLL}{rgb}{.98, .98, .98}
\definecolor{grayL}{rgb}{.9, .9, .9}
\definecolor{purpleL}{rgb}{.9735, .95, .9761}
\definecolor{purpleD}{rgb}{.8941, .8, .9043}
\definecolor{greenL}{rgb}{.9643, .9906, .9750}
\definecolor{greenD}{rgb}{.7145, .9249, .7999}
\definecolor{orangeLL}{rgb}{0.9991, 0.9846, 0.9759}
\definecolor{orangeL}{rgb}{.9982, .9692, .9518}
\definecolor{orangeD}{rgb}{.9929, .8766, .8071}

\definecolor{redL}{rgb}{1.0, 0.95, 0.95}
\definecolor{redD}{rgb}{1.0, 0.8, 0.8}
\definecolor{redDD}{rgb}{1.0, 0.4, 0.4}
\definecolor{yellowL}{rgb}{1.0, 1.0, 0.95}
\definecolor{yellowD}{rgb}{0.95, 0.95, 0.6}
\definecolor{yellowDD}{rgb}{0.8, 0.8, 0.2}
\definecolor{blueLL}{rgb}{0.98, 0.98, 1.0}
\definecolor{blueL}{rgb}{0.95, 0.95, 1.0}
\definecolor{blueD}{rgb}{0.8, 0.8, 1.0}
\definecolor{blueDD}{rgb}{0.6, 0.6, 1.0}

\definecolor{intBlueL}{RGB}{127.5, 177.5, 209.5}
\definecolor{intOrangeL}{RGB}{251,198,150}

\definecolor{HighlightColor}{rgb}{0.462745098, 0.725490196, 0.000000000}
\definecolor{nvfootlinecolor}{rgb}{0.2, 0.2, 0.2}
\definecolor{nvtextcolor}{rgb}{0.1, 0.1, 0.1}
\definecolor{NVEmerald}{rgb}{0.0, 0.521568627, 0.392156863}
\definecolor{NVAmethyst}{rgb}{0.364705882, 0.08627451, 0.509803922}
\definecolor{NVCPUBlue}{rgb}{0.0, 0.443137255, 0.77254902}
\definecolor{NVGarnet}{rgb}{0.537254902, 0.047058824, 0.345098039}
\definecolor{NVFluorite}{rgb}{0.980392157, 0.760784314, 0.0}
\definecolor{NVLightGray}{rgb}{0.803921569, 0.803921569, 0.803921569}
\definecolor{NVMediumGray}{rgb}{0.549019608, 0.549019608, 0.549019608}
\definecolor{NVDarkGray}{rgb}{0.368627451, 0.368627451, 0.368627451}
\definecolor{NVVeryLightGray}{rgb}{0.95, 0.95, 0.95}
\definecolor{NVVeryDarkGray}{rgb}{0.15, 0.15, 0.15}

\begin{document}

\title{Compact Neural Graphics Primitives with Learned Hash Probing}

\author{Towaki Takikawa}
\orcid{0000-0003-2019-1564}
\affiliation{%
  \institution{NVIDIA}
  \country{Canada}
}
\affiliation{%
  \institution{University of Toronto}
  \country{Canada}
}
\email{tovacinni@gmail.com}

\author{Thomas M\"uller}
\orcid{0000-0001-7577-755X}
\affiliation{%
  \institution{NVIDIA}
   \city{Z\"urich}
   \country{Switzerland}
  }
\email{tmueller@nvidia.com}

\author{Merlin Nimier-David}
\orcid{0000-0002-6234-3143}
\affiliation{%
  \institution{NVIDIA}
   \city{Z\"urich}
   \country{Switzerland}
  }
\email{mnimierdavid@nvidia.com}

\author{Alex Evans}
\orcid{0000-0001-7586-1735}
\affiliation{%
 \institution{NVIDIA}
  \city{London}
  \country{United Kingdom}
 }
\email{bluespoon@gmail.com}

\author{Sanja Fidler}
\orcid{0000-0003-1040-3260}
\affiliation{%
  \institution{NVIDIA}
   \city{Toronto}
   \country{Canada}
  }
\affiliation{%
  \institution{University of Toronto}
  \country{Canada}
}
\email{sfidler.com}

\author{Alec Jacobson}
\orcid{0000-0003-4603-7143}
\affiliation{%
  \institution{University of Toronto}
   \city{Toronto}
   \country{Canada}
  }
\affiliation{%
  \institution{Adobe}
   \city{Toronto}
   \country{Canada}
  }
\email{jacobson@cs.toronto.edu}

\author{Alexander Keller}
\orcid{0000-0002-9144-5982}
\affiliation{%
  \institution{NVIDIA}
   \city{Berlin}
   \country{Germany}
  }
\email{akeller@nvidia.com}

\renewcommand{\shortauthors}{Takikawa et al.}

\begin{abstract}
Neural graphics primitives are faster and achieve higher quality when their neural networks are augmented by spatial data structures that hold trainable features arranged in a grid.
However, existing feature grids either come with a large memory footprint (dense or factorized grids, trees, and hash tables) or slow performance (index learning and vector quantization).
In this paper, we show that a hash table with learned probes has neither disadvantage, resulting in a favorable combination of size and speed.
Inference is faster than unprobed hash tables at equal quality while training is only \trainingslowdown{} slower, significantly outperforming prior index learning approaches.
We arrive at this formulation by casting all feature grids into a common framework: they each correspond to a lookup function that indexes into a table of feature vectors.
In this framework, the lookup functions of existing data structures can be combined by simple arithmetic combinations of their indices, resulting in Pareto optimal compression and speed.
\end{abstract}



\begin{teaserfigure}
\vspace{-3mm}%
\hspace*{-2mm}\begin{tikzpicture}
\node (image) at (0,0) {
  \includegraphics[width=1.01\textwidth]{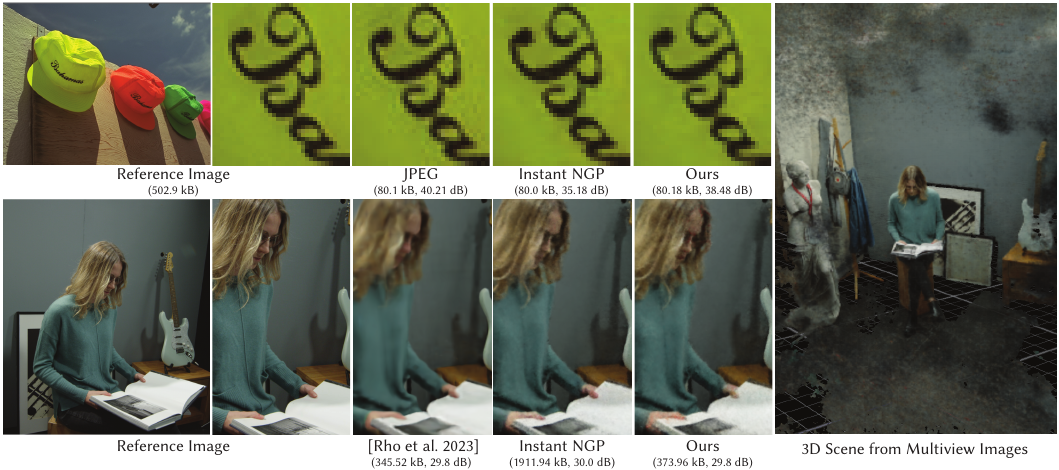}
};
\end{tikzpicture}
  \vspace{-8mm}%
  \caption{
    \emph{Compact neural graphics primitives} (Ours) have an inherently small size across a variety of use cases with automatically chosen hyperparameters.
    In contrast to similarly compressed representations like JPEG for images (top) and masked wavelet representations~\citep{rho2022masked} for NeRFs~\citep{mildenhall2020nerf} (bottom), our representation neither uses quantization nor coding, and hence can be queried without a dedicated decompression step.
    This is essential for level of detail streaming and working-memory-constrained environments such as video game texture compression.
    The compression artifacts of our method are easy on the eye: there is less ringing than in JPEG and less blur than in \citet{rho2022masked} (though more noise).
    Compact neural graphics primitives are also fast:\ training is only \trainingslowdown{} slower (depending on compression settings) and inference is faster than Instant~NGP~\citep{mueller2022instant} because our significantly reduced file size fits better into caches.\label{fig:teaser}%
  }
\end{teaserfigure}

\maketitle

\begin{figure}[H]
  \vspace{-4mm}%
  \includegraphics[width=\linewidth]{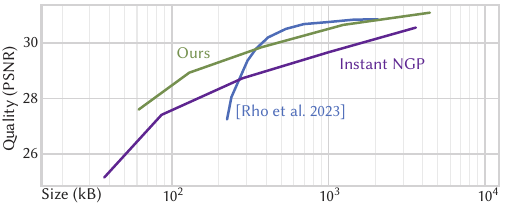}%
  \vspace{-2mm}%
  \caption{Size vs.\ PSNR Pareto curves on the NeRF scene from \autoref{fig:teaser}.
  Our work is able to outperform Instant NGP across the board and performs competitively with masked wavelet representations~\citep{rho2022masked}.\label{fig:hailey-full}}%
  \vspace{-3mm}
\end{figure}

\section{Introduction}
\label{sec:intro}

The ever increasing demand for higher fidelity immersive experiences not only adds to the bandwidth requirements of existing multimedia formats (images, video, etc.), but also fosters in the use of higher-dimensional assets such as volumetric video and light field representations.
This proliferation can be addressed by a unified compression scheme that efficiently represents both traditional and emerging multimedia content.

\emph{Neural graphics primitives (NGP)} are a promising candidate to enable the seamless integration of old and new assets across applications.
Representing images, shapes, volumetric and spatio-directional data, they facilitate novel view synthesis~(NeRFs)~\cite{mildenhall2020nerf}, generative modeling~\cite{poole2022dreamfusion, lin2022magic3d}, and light caching~\cite{muller2021real}, among more applications~\cite{neuralfields2021}.
Particularly successful are those primitives that represent data by a \emph{feature grid} that contains trained latent embeddings to be decoded by a multi-layer perceptron (MLP).
Various such feature grids have been proposed, but they usually come with a substantial memory footprint~\cite{chabra2020deep}, even when factorized into low-rank representations~\citep{Chen2022TensoRF} or represented in terms of sparse data structures~\cite{liu2020neural,yu2021plenoxels,takikawa2021neural,yu2021plenoctrees,mueller2022instant}.
In part, this limitation has been addressed by methods that \emph{learn} to index feature vectors~\cite{takikawa2022variable,li2022compressing} and leverage sparse tree structures to avoid storing feature vectors in empty space.
However, in these methods, index learning causes long training time and maintenance of sparse tree structures reduces flexibility.

Our work, \ourwork{}, combines the speed of hash tables and the compactness of index learning by employing the latter as a means of collision detection by \emph{learned probing}.
We arrive at this combination by casting all feature grids into a common framework: they all correspond to indexing functions that map into a table of feature vectors.
By simple arithmetic combinations of their indices, the data structures can be combined in novel ways that yield state-of-the-art compression vs.\ quality trade-offs.
Mathematically, such arithmetic combinations amount to assigning the various data structures to \emph{subsets} of the bits of the indexing function---thereby drastically reducing the cost of learned indexing that scales exponentially in the number of bits.

Our approach inherits the speed of hash tables while compressing much better---coming close to JPEG when representing images (\autoref{fig:teaser})---while remaining differentiable and without relying on a dedicated decompression scheme such as an entropy code.
\ourwork{} works across a wide range of user-controllable compression rates and provides streaming capabilities where partial results can be loaded in particularly bandwidth-constrained environments. %

The paper is organized as follows: we review related work and its relation to indexing schemes in \autoref{Sec:RelatedWork} before we introduce \ourwork{} in \autoref{sec:method}. We demonstrate our method in \autoref{Sec:Results} and discuss extensions, alternatives, and limitations in \autoref{Sec:Discussion} ahead of the conclusion in the last section.

\section{Related Work and Preliminaries} \label{Sec:RelatedWork}

In this article, we focus on lossy compression as it enables the highest compression rates for the multimedia under consideration.
We begin by reviewing traditional techniques before studying the connection between (neural) feature grids and indexing functions.

\subsection{Compression}

\paragraph{Traditional compression}

Lossy compression %
techniques typically employ transform coding~\cite{goyal2001theoretical} and quantization~\cite{gray1998quantization} followed by lossless entropy coding such as Huffman codes~\shortcite{huffman1952method}. On image and video content, linear transforms such as the discrete cosine~\cite{ahmed1974discrete} and wavelet~\cite{haar1909theorie} transforms are applied to exploit coefficient sparsity and reduce the visual impact of quantization errors.
Rather than transform coding, our work learns indices into a feature codebook, which is a form of vector quantization~\cite{gray1984vector, wei2000fast}, to find patterns in the data.

Texture compression relies on efficient random access to any part of the image without having to first decode the entire compressed representation.
Most methods perform block-wise compression, packing small tiles of the texture into codes of fixed size~\cite{BlockCompression,strom2005packman,beers1996rendering}.
Although our approach is different, it similarly allows for random access queries without a decompression step, enabling its potential use for texture compression in real-time renderers where feature grids have already shown promise~\citep{mueller2022instant,ntc2023}.

Volumetric compression in computer graphics~\cite{balsa2014state} similarly uses block-based coding schemes~\cite{de2016compression, tang2018real, wang2019learned, tang2020deep}.
Taking into account the often hierarchical structure of sparsity, subdivision-based spatial data structures such as trees additionally improve compression such as in OpenVDB~\cite{museth2019openvdb, museth2021nanovdb}.
By contrast, our work combines index learning and hash tables that both do not rely on a subdivision scheme for sparsity.

\paragraph{Neural compression}

In neural image compression, \emph{auto-encoder approaches} use a neural network for transform coding~\cite{theis2017lossy,balle2018variational,balle2020nonlinear}. Other works use coordinate-based neural representations to fit and compress images as continuous vector fields, some without feature grids~\cite{song2015vector, dupont2022coin++, strumpler2022implicit, lindell2021bacon} and some with feature grids~\cite{martel2021acorn,saragadam2022miner,mueller2022instant}.
Although many of these works achieve a better equal-quality compression rate than JPEG~\cite{wallace1992jpeg} at low parameter counts, high parameter counts remain challenging.
Our method is also a feature-grid and coordinate-based representation, yet performs competitively with JPEG across a wider range of qualities; see \autoref{fig:kodak-full}.

Coordinate-based neural representations are additionally applicable to volumetric and spatio-directional data; most commonly NeRFs~\citep{mildenhall2020nerf}.
Without feature grids, \citet{bird20213d} minimize the entropy of a multi-layer perceptron (MLP) and
\citet{lu2021compressive} apply vector quantization directly to the MLP parameters.
Such pure MLP methods usually have high computational cost and poor quality as compared to MLPs augmented by feature grids, so our work instead focuses on compressing feature grids while keeping the MLP sufficiently small to be fast.
\nocite{gordon2023quantizing}

\subsection{Feature Grids in the Framework of Lookup Functions}

Let us formalize feature grid methods in the following framework: they train a \emph{feature codebook} ${\codebookfeat \in \R^{\codebookfeatsize\times\featuredim}}$ of $\codebookfeatsize$ $\featuredim$-dimensional feature vectors that are associated with a conceptual grid in the $\applicationdim$-dimensional application domain.
The mapping from grid vertices ${\vertexvec = (\vertex_0, \vertex_1, \ldots) \in \Z^\applicationdim}$ to feature vectors is established by a lookup function $f(\vertexvec)$ that \emph{indexes} into the codebook, denoted by $\codebookfeat[ \cdot ]$.\footnote{Many methods maintain multiple codebooks at different resolutions, each with its own lookup function~\cite{takikawa2021neural,mueller2022instant,takikawa2022variable}, the values of which are combined before being fed to the MLP.\@
Furthermore, most methods invoke the lookup functions at several grid vertices to compute continuous outputs by interpolation~\citep{takikawa2021neural,liu2020neural}.}

\paragraph{Dense grids}

\begin{figure}
  \includegraphics[width=\columnwidth]{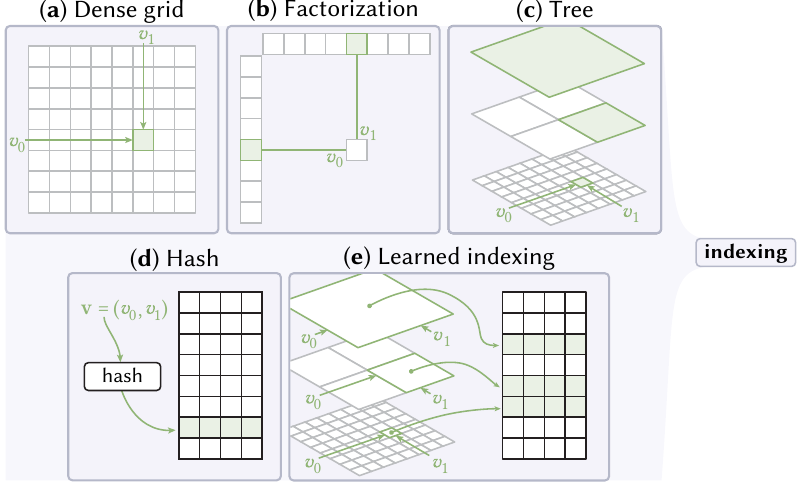}%
  \caption{
    Various indexing schemes mapping integer grid coordinates $\vertexvec = (\vertex_0, \vertex_1, \ldots)$ to feature vectors have been proposed, including
    (\textbf{a}) dense grids, %
    (\textbf{b}) $k$-planes, %
    (\textbf{c}) sparse grids and trees, %
    (\textbf{d}) spatial hashing, and %
    (\textbf{e}) learned indexing. %
    Since each scheme ultimately computes an index into a codebook of feature vectors, the schemes can be combined by arithmetic operations on the indices they produce.
    Our method combines deterministic hashing and a learned indexing as visualized in \autoref{fig:overview}.
  \label{fig:framework}
}
\end{figure}

The canonical feature grid is a dense Cartesian grid\footnote{Other tilings, such as permutohedral lattices~\citep{rosu2023permutosdf}, are also possible.}, visualized in \autoref{fig:framework}~\textbf{(a)}, that establishes a one-to-one correspondence of grid vertices to feature vectors, given for ${d = 3}$ as
\begin{align} %
  f(\vertexvec) = \codebookfeat[\vertex_0+s_0 \cdot (\vertex_1+s_1 \cdot \vertex_2)] \,,
\end{align}
where the scale ${\mathbf{s} = (s_0, s_1, \ldots)}$ defines the resolution of the grid.
Dense grids cannot adapt to sparsity in the data which makes them undesirable in practice.
For example, in 3D surface reconstruction the number of dense grid vertices is $\bigo\big(n^3\big)$ while the surfaces to be reconstructed only intersect $\bigo\big(n^2\big)$ cells.
Therefore, practitioners either combine dense grids with classic sparsification methods such as transform coding~\cite{isik2021lvac} or they choose more sophisticated indexing schemes that will be discussed next.

\paragraph{$k$-plane methods} \citep{Chen2022TensoRF,chan2021efficient,kplanes_2023,Peng2020ECCV} project the dense grid along $k$ sets of one or more axes as shown in \autoref{fig:framework}~\textbf{(b)}, and combine the resulting lower-dimensional (but still dense, usually planar) lookups arithmetically, e.g.\
\begin{align}
  f(\vertexvec) = \codebookfeat[\vertex_0 + s_0 \cdot \vertex_1]
  \cdot  \codebookfeat[s_0 \cdot s_1 + \vertex_2]
  \cdot \codebookfeat[\ldots] + \ldots \,.
\end{align}
Special cases of this scheme are equivalent to tensor decompositions of the dense grid~\citep{Chen2022TensoRF}.
While $k$-planes ensure fewer than $\bigo\big(n^\applicationdim\big)$ parameters, they makes the strong assumption that sparsity in the data can be well explained by axis aligned projections that are decoded by the MLP.\@
In practice, this is not always the case, necessitating application-specific tricks such as bounding box cropping~\citep{Chen2022TensoRF} or transform coding of the projected grids~\citep{rho2022masked} for better compression.

\begin{figure*}
  \includegraphics[width=\textwidth]{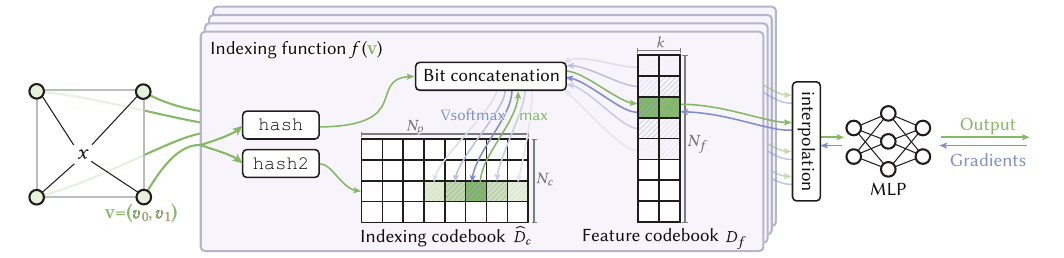}%
  \vspace{-1mm}%
  \caption{
    Overview of \ourwork{}.
    For a given input coordinate ${\pos \in \R^d}$ (far left), we find its enclosing integer grid vertices ${\vertexvec \in \Z^d}$ and apply our indexing function $f(\vertexvec)$ to each one.
    The most significant bits of the index are computed by a spatial hash (\texttt{hash}) and the least significant bits by looking up a row of $\codebookindexcoverage$ confidence values from an indexing codebook $\codebookindexwidehat$ that is in turn indexed by an auxiliary spatial hash (\texttt{hash2}), and then picking the index with maximal confidence (green arrow).
    Bitwise concatenation of the two indices yields an index for looking up from the feature codebook $\codebookfeat$, which is subsequently $d$-linearly interpolated per $\pos$ and fed into an MLP.\@
    For optimization, we propagate gradients as if the indexing codebook used a softmax instead of a hard maximum, i.e.\ we use a ``straight-through'' estimator~\citep{bengio2013estimating}.
    In practice, after each training step, we bake this ${\log_2 \codebookindexcoverage}$-bit indices of the maximal values in each row of $\codebookindexwidehat$ into an auxiliary indexing codebook $\codebookindex$ that is both compact and allows for more efficient forward evaluation of the model.
  }\label{fig:overview}
\end{figure*}

\paragraph{Spatial hashing}

Contrasting with the axis aligned parameter collisions of $k$-planes, spatial hashing~\citep{SpatialHash:03} distributes its collisions uniformly across lookups
\begin{align} \label{eq:ingp} %
  f(\vertexvec) = \codebookfeat[\texttt{hash}(\vertexvec) \bmod \codebookfeatsize] \,, \qquad \texttt{hash}(\vertexvec) = \bigoplus_{i\,=\,0}^{d-1} \, \vertex_i \cdot \pi_i \,,
\end{align}
where $\oplus$ is the binary XOR operation and $\pi_i$ are large prime numbers (optionally, ${\pi_0 = 1}$).
Well designed hash functions have the benefit that the lookups \emph{always} uniformly cover the codebook $\codebookfeat$, regardless of the underlying shape of the data, permitting sparsity to be learned independently of the data and thus application~\citep{mueller2022instant}.
But hashing also comes with the significant downside of ``scrambling'' the entries of the learned codebook $\codebookfeat$ (now a hash table), precluding structure-dependent post processing such as generative modelling or transform coding.

\paragraph{Subdivision}

Some applications~\citep{chabra2020deep,takikawa2021neural,kim2022neuralvdb,martel2021acorn} construct a sparse hierarchical data structure such as a tree whose nodes hold indices into the feature codebook:
\begin{align}
  f(\vertexvec) = \codebookfeat[\texttt{tree\_index}(\vertexvec)] \,.
\end{align}
Unfortunately, many tasks are ill-suited to such a subdivision scheme, for example image compression where subdivision heuristics are difficult to design or 3D reconstruction where sparsity is unknown a priori and only emerges during optimization~\cite{liu2020neural,yu2021plenoxels}.
Furthermore, unlike the indexing schemes above, tree traversal involves cache-unfriendly pointer chasing and therefore incurs a non-negligible performance overhead.

\paragraph{Learning the indexing function}

Rather than designing the indexing function by hand, it can also be learned from data~\citep{takikawa2022variable,li2022compressing}.
In these methods, an \emph{index codebook} ${\codebookindex \in \N^{\codebookindexsize}}$ holds the lookup indices into the feature codebook and is in turn indexed by one of the methods above.
For example, VQAD \cite{takikawa2022variable} has the lookup function
\begin{align}
  f(\vertexvec) = \codebookfeat\big[\codebookindex[\texttt{tree\_index}(\vertexvec)]\big] \,,
\end{align}
where $\codebookindex$ is trained by softmax-weighted\footnote{The softmax function $\mathbf{\sigma} : \R^d \rightarrow \R^d$ is defined as $\sigma_i(\pos) = e^{x_i} / \sum_j e^{x_j}$.} indexing into all entries of $\codebookfeat$.
This is expensive even for moderately sized feature codebooks (and prohibitive for large ones) but has no inference overhead and results in over ${10\times}$ better compression than spatial hashing.
The compression is not quite as effective as a combination of $k$-plane methods with transform coding~\citep{rho2022masked} but has the advantage that it can be cheaply queried without in-memory decompression to a larger representation.

\paragraph{Combining methods}

Using the framework of lookup functions we can relate our method to previous work: we combine learned indexing with spatial hashing by arithmetically combining their indices.
The most significant bits of our index come from Instant NGP's hash encoding~\citep{mueller2022instant} and the least significant bits are learned by a variation of VQAD~\citep{takikawa2022variable}.
Thus, our method performs \emph{learned probing} for collision resolution and information reuse in analogy to classic hash table probing methods~\cite{knuth63notes}.
This will be motivated and explained in the next section.

\section{Method}
\label{sec:method}

Our goal is to minimize the number of parameters $\theta$ and $\Phi$ of a multi-layer perceptron $m(y; \Phi)$ and its corresponding input encoding ${y = \psi(x; \theta)}$ without incurring a significant speed penalty.
Furthermore, we want to remain application agnostic and therefore avoid structural modifications such as tree subdivision and transform codings that may depend on application-specific heuristics.

Hence, we base our method on Instant NGP's multi-resolution hash encoding~\citep{mueller2022instant} and generalize its indexing function, Eq.~\eqref{eq:ingp}, by introducing learned probing.
In our lookup function, the spatial hash produces the most significant bits of the index, while the remaining user-configurable ${\log_2 \codebookindexcoverage}$ least significant bits are learned within an auxiliary index codebook ${\codebookindex \in \{0, 1, \ldots, \codebookindexcoverage-1\}^{\codebookindexsize}}$ that is in turn indexed by a second spatial hash (one that uses different prime numbers from the first).
The lookup function is illustrated in \autoref{fig:overview} and given for a single grid vertex by
\begin{align} \label{eq:our-lookup}
  f(\mathbf{v}) = \codebookfeat\big[\big(\codebookindexcoverage \cdot \texttt{hash}(\mathbf{v})\big) \bmod \codebookfeatsize + D_c[\texttt{hash2}(\mathbf{v})]\big] \,.
\end{align}
Intuitively, the index codebook $\codebookindex$, sparsified by the second spatial hash, learns to \emph{probe} the feature codebook over $\codebookindexcoverage$ values for collision resolution and information re-use.
The index codebook's size $\codebookindexsize$ as well as its probing range $\codebookindexcoverage$ are hyperparameters of our method that extend those inherited from Instant NGP; see \autoref{tab:params}.

\begin{table}
  \small
  \caption{Hyperparameters of our method and recommended ranges. We inherit most parameters from Instant NGP~\citep{mueller2022instant} and introduce two additional ones pertaining to the index codebook. {\color{gray}Gray parameters} are unaffected by our method and therefore set to the same values as in Instant NGP; the choice of remaining parameters is explained in \autoref{sec:method}.}%
  \newcolumntype{R}{>{\raggedleft\arraybackslash}X}%
  \label{tab:params}%
  \vspace{-3mm}%
  \begin{tabularx}{\linewidth}{cllr}%
    \toprule%
    \textit{Source} & \textit{Parameter} & \textit{Symbol} & \textit{Value} \\
    \midrule%
    \multirow{2}{*}{\shortstack[c]{new in\\our method}} & Index probing range & $\codebookindexcoverage$ & $2^1$ to $2^4$ \\
    & Index codebook size &
    $\codebookindexsize$ & $2^{10}$ to $2^{24}$ \\%[1.5mm]
    \midrule
    \multirow{5}{*}{\shortstack[c]{inherited from\\Instant NGP}} & Feature codebook size & $\codebookfeatsize$ & $2^{6}$ to $2^{12}$ \\
    & \color{gray} Feature dimensionality & \color{gray} $\featuredim$ & \color{gray} $2$ \\
    & \color{gray} Number of levels & \color{gray} $\levels$ & \color{gray} 16 \\
    & \color{gray} Coarsest resolution & \color{gray} $\minresolution$ & \color{gray} $16$ \\
    & \color{gray} Finest resolution & \color{gray} $\maxresolution$ & \color{gray} $512$~to~$524288$ \\
    & \color{gray} Num.\ hidden neurons & \color{gray} $N_\text{neurons}$ & \color{gray} $64$ \\
    \bottomrule%
  \end{tabularx}%
  \vspace{-2mm}
\end{table}

\begin{figure*}
  \vspace{-3mm}%
  \includegraphics[width=\textwidth]{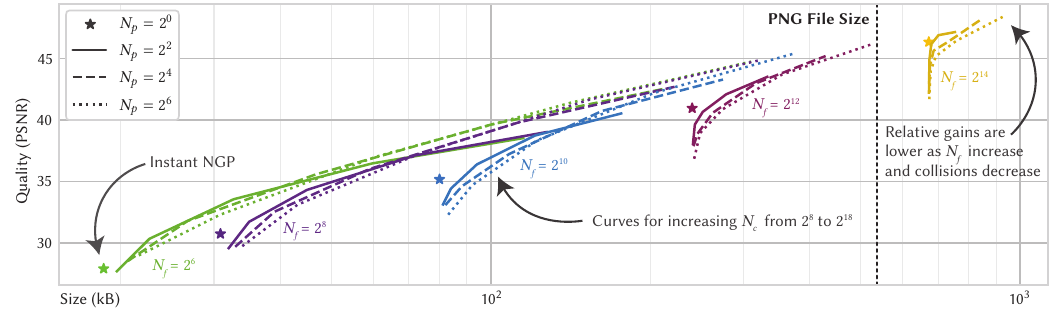}%
  \vspace{-2mm}%
  \caption{PSNR vs.\ file size for varying hyperparameters in compressing the Kodak image dataset. We sweep three parameters: the feature codebook size $\codebookfeatsize$ (colors), the index codebook size $\codebookindexsize$ (curves ranging from ${2^{12}}$ to ${2^{20}}$), and the probing range $\codebookindexcoverage$ (dashing and dotting).
  A value of ${\codebookindexcoverage = 1}$ corresponds to Instant NGP (shown as $\star$) and has no curve because it is invariant under $\codebookindexsize$.
  We see that the optimal curve at a given file size $N$ has a feature codebook size (same-colored $\star$) of roughly ${\codebookfeatsize = \nicefrac{1}{3}N}$ and index codebook size ${\codebookindexsize = \nicefrac{2}{3}N}$.
  Small probing ranges (solid curves) are sufficient for good compression---in-fact optimal for small values of $\codebookindexsize$ (left side of curves)---but larger probing ranges (dashed and dotted curves) yield further small improvements for large values of $\codebookindexsize$ (right side of curves) at the cost of increased training time.\label{fig:ablation-image}}%
  \vspace{-2mm}
\end{figure*}

Following \citet{takikawa2022variable}, we maintain two versions of the index codebook: one for training ${\codebookindexwidehat \in \R^{\codebookindexsize\times\codebookindexcoverage}}$ that holds confidence values for each of the $\codebookindexcoverage$ features in the probing range, and one for inference ${\codebookindex \in \{0, 1, \ldots, \codebookindexcoverage-1\}^{\codebookindexsize}}$ that holds ${\log_2 \codebookindexcoverage}$-bit integer indices corresponding to the probe offset with largest confidence.
Compared to Instant NGP, the only inference-time overhead is the index lookup from $\codebookindex$.
Furthermore, our smaller parameter count leads to improved cache utilization; we hence achieve similar and in some cases better inference performance as shown in \autoref{tab:performance}.

\begin{figure}
  \vspace{-1mm}%
  \includegraphics[width=\linewidth]{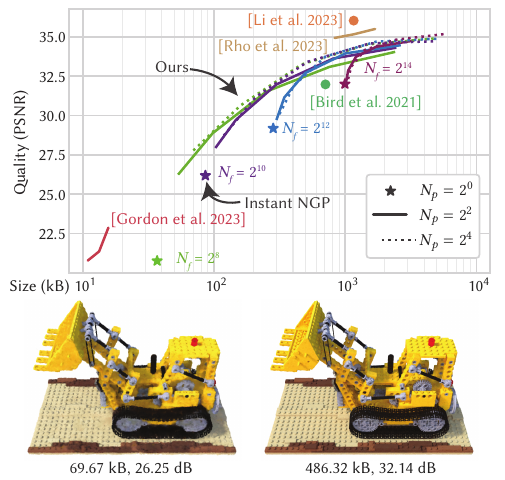}%
  \vspace{-2mm}%
  \caption{PSNR vs.\ file size for varying hyperparameters in compressing the NeRF Lego digger. The layout is the same as \autoref{fig:ablation-image}. We also show rendered images of our compressed representation at two quality settings.}
  \vspace{-3mm}%
\label{fig:ablation-nerf}
\end{figure}

\begin{table}
  \centering
  \caption{Training and inference time overheads of \ourwork{}. Training times are measured for an iteration of training on the NeRF Lego digger dataset. Inference times are for $2^{18}$ lookups on a single multiresolution level. The relative training overhead (denoted with $n\times$) is measured with respect to Instant NGP (${\codebookfeatsize = 2^{16}}$), ranging from \trainingslowdown{}.
  The largest impact on speed has the probing range $\codebookindexcoverage$, whereas $\codebookindexsize$ (shown) and $\codebookfeatsize$ (see \citet{mueller2022instant}) only have a weak effect.\label{tab:performance}}%
  \vspace{-3mm}%
   \resizebox{\linewidth}{!}{
  \begin{tabular}{lccccccc}
  \toprule
  \textit{Method} & $N_f$ & $N_c$ & $N_p$ & \thead{\textit{Training time} \\ \textit{per iteration}} & \thead{\textit{Inference time} \\ \textit{for $2^{18}$ lookups}} & \thead{\textit{Quality} \\ \textit{(PSNR dB)}}\\
  \midrule
  \multirow{3}{*}{I~NGP} & $2^{16}$ & n/a & $2^0$ & 5.4~ms & 28.7$\mu$s & 33.60~dB \\
   & $2^{14}$ & n/a & $2^0$ & 5.1~ms & $13.7\mu$s & 32.00~dB\\
  & $2^{8}$ & n/a & $2^0$ & 4.5~ms & 9.8$\mu$s & 19.04~dB\\
  \midrule
  \multirow{6}{*}{Ours} & $2^{8}$ & $2^{12}$ & $2^2$ & 6.8ms (1.26$\times$) & 10.1$\mu$s & 26.25~dB \\
  & $2^{8}$ & $2^{16}$ & $2^2$ & 6.8~ms (1.26$\times$) & 10.1$\mu$s & 31.58~dB \\
  & $2^{8}$ & $2^{12}$ & $2^3$ & 8.3~ms (1.53$\times$) & 10.1$\mu$s & 27.13~dB \\
  & $2^{8}$ & $2^{16}$ & $2^3$ & 8.5~ms (1.57$\times$) & 10.2$\mu$s & 32.58~dB\\
  & $2^{8}$ & $2^{12}$ & $2^4$ & 12.7~ms (2.35$\times$) & 10.2$\mu$s & 27.67~dB\\
  & $2^{8}$ & $2^{16}$ & $2^4$ & 14.1~ms (2.61$\times$) & 10.2$\mu$s & 33.24~dB\\
  \bottomrule
  \end{tabular}}%
  \vspace{-4mm}
\end{table}

\paragraph{Training}

In the forward pass we use $\codebookindex$ to look up the feature with largest confidence and in the backward pass we distribute gradients into \emph{all} features within the probing range, weighted by the softmax of their confidence values from $\codebookindexwidehat$ (see \autoref{fig:overview}).
This strategy of combining a discrete decision in the forward pass with continuous gradients in the backward pass is also known as a ``straight-through'' estimator that helps to learn hard non-linearities~\citep{bengio2013estimating}.

By keeping the learned number of bits $\log_2 \codebookindexcoverage $ small, we limit the number of features and confidence values that need to be loaded in the backward pass.
And since the learned bits are the least significant ones, their corresponding features lie adjacent in memory, usually located in the same cache line and thereby incurring only a moderate training overhead of \trainingslowdown{} (see \autoref{tab:performance}) while realizing compression rates on par with the orders of magnitude slower VQAD~\citep{takikawa2022variable}.

\paragraph{Selecting hyperparameters}

Recall that our method inherits its hyperparameters from Instant~NGP and introduces two new ones: the index codebook size $\codebookindexsize$ and its probing range $\codebookindexcoverage$; see \autoref{tab:params} for a complete list.
To find quality-maximizing parameters, we recommend the following scheme inspired by Figures~\ref{fig:ablation-image} and \ref{fig:ablation-nerf}, which we use in all our following results.
First, set ${\codebookindexsize = 1}$ and ${\codebookindexcoverage = 1}$, turning the method into Instant NGP as indicated by $\star$ in the figure.\@
Second, set the feature codebook size $\codebookfeatsize$ according to the desired lower bound on the compressed size.
Third, double $\codebookindexsize$ until a reasonable maximum value (usually $\codebookindexsize=2^{16}$).
Lastly, if even higher quality is desired, double $\codebookfeatsize$. The remaining parameter $\codebookindexcoverage$ can be tuned to taste, as this parameter governs how expensive the training is, but a higher value tends to produce slightly better Pareto tradeoffs between size and quality.

\section{Results} \label{Sec:Results}

We have implemented our algorithm on top of the version of Instant~NGP in the PyTorch-based Kaolin Wisp library~\cite{takikawa2022kaolin}.
Computationally expensive operations like sparse grid ray tracing and feature grid lookups of both Instant~NGP and our method are accelerated by custom CUDA kernels called from  PyTorch. All results are measured on an NVIDIA RTX 6000 Ada GPU.

\paragraph{Performance.}

\autoref{tab:performance} lists inference and training times of our method on the NeRF Lego digger from \autoref{fig:ablation-nerf}.
Compared to Instant~NGP, our \trainingslowdown{} training overhead scales with the probing range $\codebookindexcoverage$, confirming the analysis in \autoref{sec:method} and exposing a trade-off between training speed and compression to the user.
Since the compression benefit of larger probing ranges quickly falls off, we cap ${\codebookindexcoverage \leq 2^4}$ in all our experiments, manifesting the worst-case overhead of $2.6\times$.
An important performance consideration for training is the accumulation of gradients into the feature codebook $\codebookfeat$.
Since our method uses very small codebooks ${\codebookfeatsize \in [2^6, 2^{12}]}$, special care must be taken on massively parallel processors, such as GPUs, to first accumulate gradients in threadblock-local memory before broadcasting them into RAM.\@
This avoids contention that would otherwise make training ${\sim\!7\!\times}$ slower.

\autoref{tab:performance} also demonstrates that \ourwork{} has \emph{faster} inference than Instant~NGP at roughly equal quality settings.
This is because our method has a much smaller size (${\codebookfeatsize = 2^{16}}$ vs.\ ${\codebookfeatsize = 2^8, \codebookindexsize = 2^{16}}$) and thereby fits better into caches.
The only inference overhead of our method is the additional index lookup from $\codebookindex$,
which we find negligible (0.4$\mu$s at ${\codebookfeatsize = 2^8}$).

\paragraph{Image compression.}

\begin{figure}
  \includegraphics[width=\linewidth]{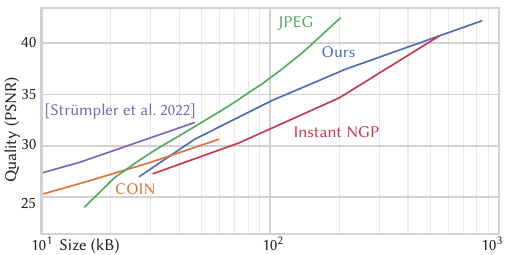}%
  \caption{PSNR vs.\ file size on the Kodak image dataset using parameters ${N_f=2^6}$ and ${N_p=2^4}$ and varying $N_c$ (blue curve ranging from ${2^{12}}$ to ${2^{20}}$). On this dataset, our method performs close to JPEG at small file sizes and worse at larger ones. At small file sizes, our representation is dominated by floating point parameters like the MLP and the feature codebook.
  Competing methods that quantize pure MLPs perform better in this regime~\citep{strumpler2022implicit,dupont2021coin}, whereas we omit quantization for simplicity and flexibility.
  At visually pleasant targets ($\sim$35dB and above) these prior works do not scale as it is difficult to train pure MLPs to such qualities.\label{fig:kodak-full}}%
  \vspace{-2mm}%
\end{figure}

\begin{figure}
  \includegraphics[width=\linewidth]{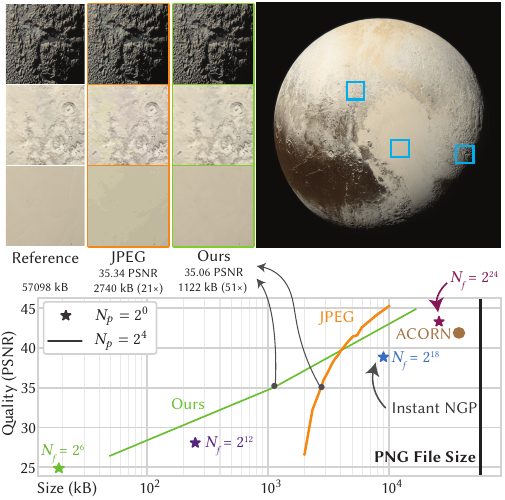}%
  \vspace{-2mm}%
  \caption{We fit Compact NGP to the ${8000\!\times\!8000}$px Pluto image using parameters $N_f=2^6$ and $N_p=2^4$ and varying $N_c$ (green curve ranging from ${2^{12}}$ to ${2^{24}}$). We show that we are able to outperform JPEG on a wide range of quality levels.
  The qualitative comparisons at \textit{equal size} (insets) show the visual artifacts exhibited by different methods: while JPEG has color quantization arfitacts, ours appears slightly blurred.\label{fig:pluto}}%
  \vspace{-2mm}
\end{figure}

\autoref{fig:kodak-full} shows the quality vs.\ size tradeoff of our method on the Kodak image dataset, which consists of 24 images of ${768\!\times\!512}$ pixels.
The figure also shows JPEG as well as prior coordinate MLP methods.
On this dataset, our method performs close to JPEG at small file sizes and worse at larger ones.
At small file sizes, our representation is dominated by floating point parameters like the MLP and the feature codebook, causing competing methods that apply quantization on top of pure MLPs~\citep{strumpler2022implicit,dupont2021coin} to compress better.
However, these methods do not scale to higher quality targets ($\sim$35dB and above) as it is difficult to train pure MLPs to such qualities.
To demonstrate the better scaling of our method, we investigate a much larger ${8000\!\times\!8000}$ image of Pluto in \autoref{fig:pluto} on which we outperform both JPEG on most practical sizes ($\sim$megabyte) and prior neural large-scale methods (Instant NGP~\citep{mueller2022instant} and ACORN~\citep{martel2021acorn}) at high quality settings.
Our method is also evaluated against texture compression methods in \autoref{fig:table-texture-compression}.

\paragraph{NeRF compression}

We evaluate NeRF compression on a real-world scene in Figures~\ref{fig:teaser}~and~\ref{fig:hailey-full} as well as synthetic scenes~\citep{mildenhall2020nerf} in \autoref{fig:ablation-nerf} (Lego) and \autoref{fig:table-nerf} (full dataset).
We compare with several contemporary NeRF compression techniques that are mostly based on TensoRF~\cite{Chen2022TensoRF}.
We report numbers from the original papers where available.
For the real world scene, we ran
masked wavelets~\cite{rho2022masked} as a strong and recent baseline.
In both scenes, we outperform Instant~NGP in terms of quality vs.\ size.
On the synthetic scene (\autoref{fig:ablation-nerf}), our Pareto front lies slightly below the specialized baselines that use scalar quantization and coding, and in the real-world scene our Pareto front is competitive (\autoref{fig:hailey-full}) despite our work requiring neither.

The zoom-ins in \autoref{fig:teaser} reveal distinct visual artifacts of the different methods, even though their PSNR is the same.\@
Masked wavelets~\cite{rho2022masked} produce blurrier results whereas \ourwork{} yields a sharper reconstruction with high frequency noise similar to that of Instant~NGP.\@

\begin{figure}
  \includegraphics[width=\linewidth]{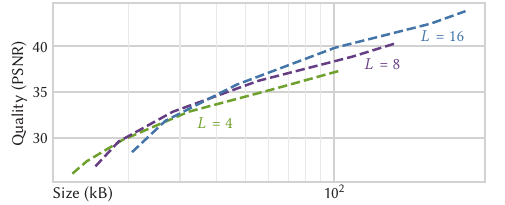}%
  \vspace{-2mm}%
  \caption{Impact of the number of multiresolution levels $L$ on PSNR vs.\ size.
  We use the parameters ${\codebookfeatsize=2^6}$ and ${\codebookindexcoverage=2^4}$ while varying $\codebookindexsize$ (curve ranging from ${2^{12}}$ to ${2^{20}}$) and $L$ on the image compression task from \autoref{fig:teaser}.
  The default value ${L=16}$ (inherited from Instant NGP) performs well for a large range of sizes, particularly in the hundreds of kB range that is most practical.
  Yet, a lower number of levels results in a better Pareto curve at smaller sizes that could be used if one wanted to compete with MLP based image compression techniques; cf.\ \autoref{fig:kodak-full}.\label{fig:numlevels}}%
  \vspace{-1mm}
\end{figure}

\begin{figure}
  \includegraphics[width=\linewidth]{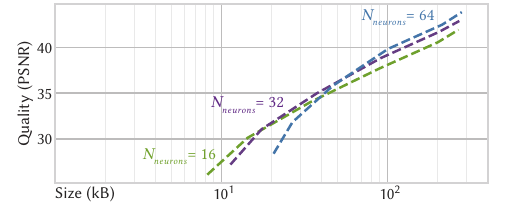}%
  \vspace{-2mm}%
  \caption{Impact of the MLP width ${N_\text{neurons}}$ on PSNR vs.\ size.
  The parameter sweeps over $\codebookfeatsize$, $\codebookindexcoverage$, and $\codebookindexsize$ are the same as \autoref{fig:numlevels}.
  A similar conclusion can be drawn: the default value ${N_\text{neurons}=64}$ (inherited from Instant NGP) performs well at practical sizes, whereas a better Pareto front can be achieved at smaller sizes.\label{fig:hiddendim}}%
  \vspace{-3mm}
\end{figure}

\paragraph{Additional hyperparameter ablations}

Aside from the feature codebook size $\codebookfeatsize$, we inherit the default hyperparameters of Instant~NGP for a apples-to-apples comparisons.
To verify that these defaults are reasonable, we sweep the number of multiresolution levels $L$ in \autoref{fig:numlevels} and the number of hidden neurons $N_\text{neurons}$ in \autoref{fig:hiddendim}.
The default values ${L=16}$ and ${N_\text{neurons}=64}$ perform well for a large range of sizes, particularly in the hundreds of kB range that is most practical.
Yet, lower values produce better Pareto frontiers at very small file sizes that could be used if one wanted to compete with MLP based image compression techniques; cf.\ \autoref{fig:kodak-full}.
However, we believe that the hundreds of kB range is more relevant in practice and we therefore stick to the default values for simplicity.

\section{Discussion and Future Work} \label{Sec:Discussion}

\ourwork{} has been designed with content distribution in mind where the compression overhead is amortized and decoding on user equipment must be low cost, low power, and multi-scale for graceful degradation in bandwidth-constrained environments.
As an example, NeRFs may be broadcasted and decoded
on large numbers of end-user devices, possibly in real-time to enable live streaming video NeRFs.
More generally, (learnable) compression codecs will enable the next generation of immersive content of which live streaming of NeRFs are just an example and other applications, like video game texture compression and volumetric video, being right around the corner.

\paragraph{Quality and compression artifacts.}
Beyond measuring PSNR, it is worth studying the qualitative appearance of compression artifacts with our method.
Compared to JPEG, our method appears to produce less ringing at the cost of a small amount of additional blur, whereas in NeRF our methods looks similar to Instant NGP: sharp, but with high-frequency noise.
This is in contrast to \citet{rho2022masked}, who produce a smoother yet blurry reconstruction; see \autoref{fig:teaser}.
Since we measure error in terms of PSNR, which is based on the $\mathcal{L}_2$ error, blurry results yield lower error than the human visual system might expect~\citep{zhao2016loss}.

\paragraph{From \texttt{float} to \texttt{int}.}
Our method shifts the storage cost from being \texttt{float}-dominated to \texttt{int}-dominated.
In the settings we test in, we see that this tradeoff is favorable, particularly because our integers have only ${\log_2 \codebookindexcoverage}$ bits---many fewer than than even 16-bit half precision floats.
We have additionally investigated several methods that reduce the entropy of our learned indices (e.g.\ through additional terms in the loss), coupled to entropy coding, but so far with little success that does not warrant forfeiture of random access lookups.
Alternatively, data-adaptive quantization of floats may reduce the bit count further than using an index codebook, but better training strategies are required to this end.
We believe that further research into data-adaptive \texttt{float} quantization as well as \texttt{int} entropy minimization will be fruitful.

\paragraph{Entropy coding.}
Our method was inspired by a method that has spatial locality built-in~\citep{takikawa2022variable} (i.e.\ the index codebook represented by a tree).
Such spatial locality could be exploited by an entropy coder much better than the spatial hash table that we use.
We chose spatial hashing for being agnostic of the application~\citep{mueller2022instant}---and it performs competitively with transform and entropy coded prior work nonetheless---but if future research could devise local data structures that have the same flexibility and performance as hash tables, it will likely be worthwhile to utilize those instead of hashing.

\paragraph{Alternatives to straight-through estimators}
In our work we use the softmax function along with the straight-through estimator to learn indexing. While effective, this can be computationally expensive for large indexing ranges as this requires backpropagation on all possible indices. As such, it may be worthwhile to explore the various sparse~\cite{Sparsemax, peters-etal-2019-sparse, SparseAltSoftmax} and stochastic~\cite{SparseSoftmax, dropmax} variants have been proposed in the literature. Proximity-based indexing such as locality-sensitive hashing and the nearest-neighbour queries used in VQ-VAE~\cite{van2017neural} may be relevant as well.

\section{Conclusion}

We
propose to view feature grids and their associated neural graphics primitives through a common lens: a unifying framework of lookup functions.
Within this framework it becomes simple to mix methods in novel ways, such as our \ourwork{} that augments efficient hash table lookups with low-overhead learned probing.
The result is a state-of-the-art combination of compression and performance while remaining agnostic to the graphics application in question.
\ourwork{} has been designed with real-world use cases in mind where random access decompression, level of detail streaming, and high performance are all crucial (both in training and inference).
As such, we are eager to investigate its future use in streaming applications, video game texture compression, live-training as in radiance caching, and many more.

\begin{acks}
The Lego Bulldozer scene of \autoref{fig:ablation-nerf} was created by \href{https://www.blendswap.com/blend/11490}{Blendswap user Heinzelnisse}.
The Pluto image of \autoref{fig:pluto} was created by NASA/Johns Hopkins University Applied Physics Laboratory/Southwest Research Institute/Alex Parker.
We thank David Luebke, Karthik Vaidyanathan, and Marco Salvi for useful discussions throughout the project.
\end{acks}

\bibliographystyle{ACM-Reference-Format}
\bibliography{main}

\begin{table*}
\centering
  \caption{Quantiative results on the full synthetic dataset from~\citet{mildenhall2020nerf}, showing a near-quality (PSNR) comparison between Instant~NGP and our work. We see that we are able to achieve similar quality across the entire dataset with a $2.8\times$ more compact representation.\label{fig:table-nerf}}%
\begin{tabular}{lccccccccccccc}
\toprule
\textit{Method} & $N_f$ & $N_c$  & $N_p$ &  \textit{Mic} & \textit{Ficus} & \textit{Chair} & \textit{Hotdog} & \textit{Materials} & \textit{Drums} & \textit{Ship} & \textit{Lego} & \textit{avg.} & \textit{Size (kB)} \\
\midrule
I~NGP & $2^{14}$ & n/a & $2^0$ & $35.08$ & $30.99$ & $32.59$ & $34.99$ & $28.73$ & $25.36$ & $27.71$ & $32.03$ & $30.93$ & $1000$~kB\\
\midrule
Ours & $2^{8}$ & $2^3$ & $2^{16}$ & $33.88$ & $32.08$ & $32.05$ & $34.26$ & $28.32$ & $24.71$ & $27.71$ & $32.31$ & $30.66$ & $357$~kB \\
\bottomrule

\end{tabular}
\end{table*}

\def\checkmark{\tikz\fill[scale=0.4](0,.35) -- (.25,0) -- (1,.7) -- (.25,.15) -- cycle;}
\begin{table*}
\centering
  \caption{Quantiative results on texture compression on the \textit{Paving Stones} texture set, retrieved from \texttt{https://ambientcg.com}, showing the tradeoff between quality (PSNR) and size (kB) for different methods. We compare against traditional texture compression baselines (BC) as well as recent neural baselines (NTC~\cite{ntc2023}). We borrow the results from \citet{ntc2023}. Although our work does not outperform NTC, which uses a specialized architecture for textures with quantization, we are still able to outperform BC and Instant~NGP at similar size. We only report average across all channels for BC as that was the only data available, and compare against the NTC results without mipmaps (which increase quality) for fair comparison.\label{fig:table-texture-compression}}%
\begin{tabular}{lccccccccccc}
\toprule
\textit{Method} & \textit{Quantization} & $N_f$ & $N_c$  & $N_p$ &  \textit{Diffuse} & \textit{Normal} & \textit{Roughness} & \textit{AO} & \textit{Displacement} & \textit{avg.} & \textit{Size (kB)}  \\
\midrule

I~NGP & & $2^{16}$ & n/a  & n/a & $21.58$ & $22.32$ & $26.79$ & $27.72$ & $35.62$ & $24.75$ & $3761$~kB \\
I~NGP & & $2^{14}$ & n/a & n/a & $19.91$ & $20.51$ & $26.61$ & $25.56$ & $30.07$ & $22.61$ & $1049$~kB \\
\midrule
BC & \checkmark & n/a & n/a & n/a & n/a & n/a & n/a & n/a & n/a & 23.25 & $3500$~kB \\
\midrule
NTC & \checkmark & n/a & n/a & n/a & 26.10 & 27.17 & 29.37 & 31.22 & 40.59 & $29.00$ & $3360$~kB \\
\midrule
Ours &  & $2^{10}$ & $2^{20}$& $2^3$ & $24.02$ & $25.00$ & $27.90$ & $29.94$ & $36.18$ & $26.69$ & $3494$~kB \\
Ours &  & $2^8$ & $2^{18}$& $2^3$ & $21.55$ & $22.61$ & $26.94$ & $27.43$ & $33.74$ & $24.51$ & $1173$~kB \\
\bottomrule

\end{tabular}
\end{table*}

\end{document}